\pdfoutput=1

\documentclass[11pt]{article}

\usepackage{EMNLP2023}

\usepackage{times}
\usepackage{latexsym}
\usepackage[T1]{fontenc}
\usepackage[utf8]{inputenc}
\usepackage{amsfonts}       
\usepackage{microtype}
\usepackage{multirow}
\usepackage{subfigure}
\usepackage{inconsolata}
\usepackage{graphicx} 
\usepackage{booktabs}
\usepackage{bbding}

\def\logo{\makebox[0pt][l]{\hspace{0pt}\raisebox{-0.3ex}{\includegraphics[height=24pt]{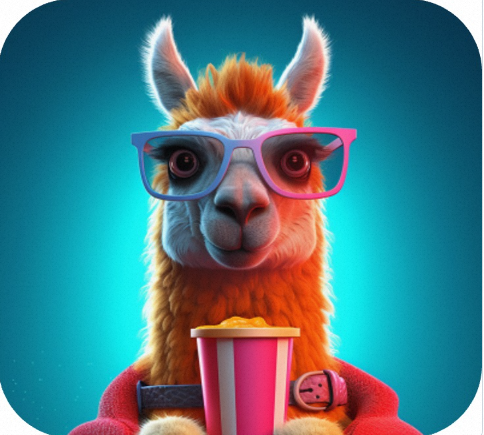}}}}
\title{\logo \ \ \ \ \ \ \ \   \  Video-LLaMA 

An Instruction-tuned Audio-Visual Language Model for Video Understanding}

\author{
Hang Zhang\textsuperscript{\rm 1 \rm2}  \ \ \ \ \ \ \ \ 
Xin Li\textsuperscript{\rm 1  \rm2}\thanks{~ Xin Li is the corresponding author.} \ \ \ \ \ \ \ \ 
Lidong Bing\textsuperscript{\rm1 \rm 2} \\
\textsuperscript{\rm 1} 
DAMO Academy, Alibaba Group \\
\textsuperscript{\rm 2}
Hupan Lab, 310023, Hangzhou, China \\
\tt{\{zh401075, xinting.lx, l.bing\}@alibaba-inc.com} 
}

\begin{document}
\maketitle
\begin{abstract}
We present Video-LLaMA\footnote{The video demonstration is available at \url{https://youtu.be/RDNYs3Rswhc}} a multi-modal framework that empowers Large Language Models (LLMs) with the capability of understanding both visual and auditory content in the video. Video-LLaMA bootstraps cross-modal training from the frozen pre-trained visual \& audio encoders and the frozen LLMs. Unlike previous works that complement LLMs to process the visual or audio signals only~\citep{zhu2023minigpt,liu2023visualit,huang2023audiogpt}, Video-LLaMA enables video comprehension by tackling two challenges: (1) capturing the temporal changes in visual scenes, (2) integrating audio-visual signals. To counter the first challenge, we propose a Video Q-former to assemble a pre-trained image encoder into our video encoder and introduce a video-to-text generation task to learn video-language correspondence. For the second challenge, we leverage ImageBind~\citep{girdhar2023imagebind}, a universal embedding model aligning multiple modalities, as the pre-trained audio encoder and introduce an Audio Q-former on top of ImageBind to learn reasonable auditory query embeddings for the LLM module. To align the output of both visual \& audio encoders with LLM's embedding space, we first train Video-LLaMA on massive video/image-caption pairs and then tune our model with visual-instruction datasets of moderate amount but higher quality. We found Video-LLaMA shows the ability to perceive and comprehend video content and generate meaningful responses grounded in the visual and auditory information presented in the videos.  
\end{abstract}

\section{Introduction}

Large Language Models (LLMs)~\citep{chowdhery2022palm,bai2022constitutional,openai2023gpt4tr} have demonstrated remarkable capability of understanding and following user intentions and instructions\footnote{https://chat.openai.com/chat}\footnote{https://www.anthropic.com/product}\footnote{https://bard.google.com/}. Typically, the user requests and the corresponding responses from LLMs are all in texts, however, text-only human-computer interaction is not sufficient for many application scenarios because real-world information is usually multi-modal. In order to further explore the potential of LLMs, many researchers attempt to endow LLMs with the capability of understanding multi-modal content~\citep{huang2023audiogpt,zhang2023speechgpt,yin2023survey}.

Among these efforts, \citet{alayrac2022flamingoav,wang2022ofa,huang2023languagein,xu2023mplug2,zhang2023speechgpt,sun2023generative} pre-train multi-modal LLMs with massive interleaved image-text data or speech-text data to accommodate multi-modal input. Meanwhile, another group of works adopts a more parameter-efficient way by complementing LLMs with off-the-shelf vision or speech foundation models to achieve multi-modal understanding ~\citep{li2023blip2bl,zhu2023minigpt,liu2023visualit,ye2023mplugowl,zhang2023vpgtrans,huang2023audiogpt,wu2023onda,pandagpt,li2023otter}. 

Despite their effectiveness, these approaches are dedicated to aligning the input from exactly one additional modality with text (i.e., image or audio), which is unsatisfactory for video understanding. Concretely, empowering LLMs to understand video requires comprehensive processing  for different modalities including visual input, auditory input, and textual output, which is more challenging than image-only understanding and audio-only understanding tasks.
Although there are several recent works attempt to unleash the video understanding capability of LLMs~\citep{li2023videochatcv,maaz2023videochatgpt,luo2023valley}, their primary objective is to comprehend only the visual content of the video, with the auditory content remaining unused.


\begin{table}[t!]
\centering
\resizebox{0.47\textwidth}{!}{
\begin{tabular}{|l|ccc|} \toprule
\multirow{2}{*}{Model Name} & \multicolumn{3}{c|}{Ability}                                                 \\
                            & Static Image & Silent Video & Audio   \\ \midrule
BLIP2~\cite{li2023blip2bl}  &   \Checkmark  &                &                  \\
MiniGPT4~\cite{zhu2023minigpt}                    &   \Checkmark  &                &                   \\
LLaVA~\cite{liu2023visualit}           &   \Checkmark  &                &               \\
mPLUG-Owl~\cite{ye2023mplugowl}                  &   \Checkmark  &   \Checkmark   &              \\
VideoChat~\cite{li2023videochatcv}     &   \Checkmark  &   \Checkmark   &               \\ 
AudioGPT~\cite{huang2023audiogpt} &    &      &     \Checkmark  \\ 
Video-ChatGPT~\cite{maaz2023videochatgpt}     &   \Checkmark  &   \Checkmark   &               \\ \midrule
Video-LLaMA                  &   \Checkmark    &  \Checkmark   &    \Checkmark   \\ \bottomrule
\end{tabular}
}
\caption{Comparison with popular multi-modal large language models. Video-LLaMA has the unique ability to comprehend auditory and visual information simultaneously.}
\label{tab.comparison}
\end{table}

\begin{figure*}[ht]
    \centering
    \includegraphics[scale=0.44]{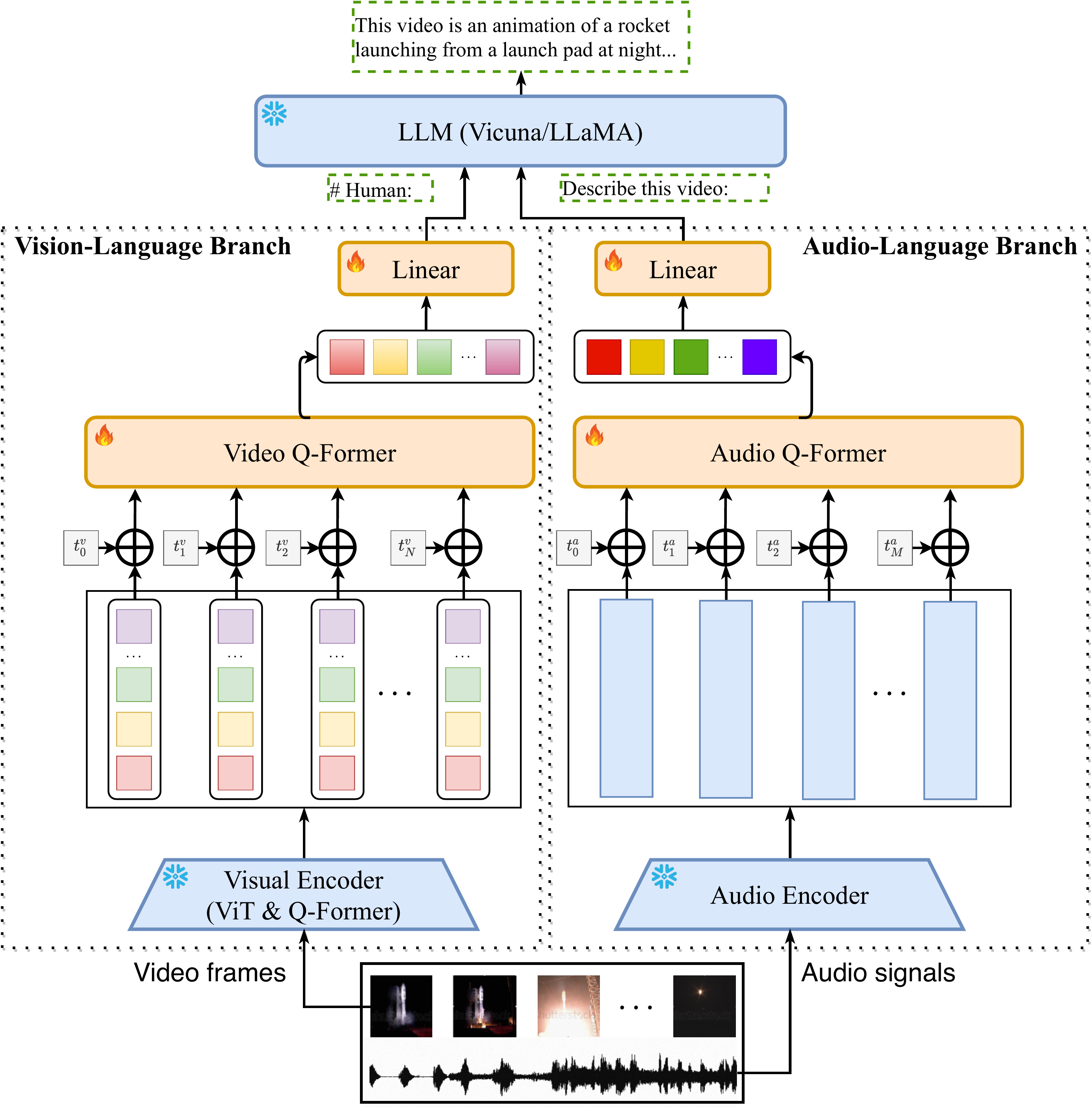}
    \caption{Overall architecture of Video-LLaMA.}
    \label{fig:architecture}
\end{figure*}

In this work, to fill in the blank of audio-visual LLMs, we investigate the possibility of building multi-modal LLMs that support the input of video and allow users to chat with computers around the user-uploaded video, which is usually composed of multiple video frames and audio. Instead of employing external perception models to convert visual/auditory signals to textual signals~\citep{shen2023hugginggpt,li2023videochatcv}, we choose to build an end-to-end model that can handle the data from multiple modalities within one single framework. Specifically, we adopt the idea of BLIP-2~\citep{li2023blip2bl} to guarantee the efficiency of cross-modal pre-training. To explicitly capture the change of visual scenes in the video, we use a pre-trained visual encoder to separately compute frame representations. Then, we introduce a frame embedding layer to inject temporal information and a video Q-Former to generate visual query tokens. As for the audio signals from the video, we additionally leverage a pre-trained audio encoder as well as an audio Q-former to learn reasonable auditory query embeddings (see the right part of Figure~\ref{fig:architecture}). 


To align textual output with video, we devise multi-branch cross-modal pre-training to learn the vision-language correspondence and the audio-language correspondence. For vision-language correspondence, we first pre-train the vision-related components on a large-scale video caption dataset with a video-clips-to-text generation task. To enhance the understanding of static visual concepts, we also add image-caption data into this pre-training stage. Then, we further fine-tune these components on a video-based conversation dataset to execute visual instruction tuning. For the alignment between the audio encoder and language decoder, we further pre-train the audio-related components on an audio caption dataset with an audio-to-text generation task. For the audio-language correspondence, we leverage Imagebind~\citep{girdhar2023imagebind} as an encoder, which performs exceptionally well in aligning different modalities to a common embedding space. Given the limited availability of audio-text data, we also utilize vision-text data to train the audio-related components. These components learn to align the common embedding space provided by Imagebind with the embedding space of LLMs. Despite not being explicitly trained with audio-text data, Video-LLaMA exhibits a remarkable zero-shot audio understanding capability during inference. 

As shown in Table~\ref{tab.comparison}, our Video-LLaMA stands out from other existing multi-modal LLMs in terms of its distinctively comprehensive comprehension of audiovisual modal information in videos. In summary, our contributions are as follows:

\indent $\bullet$ We propose Video-LLaMA, a multi-modal framework that enables LLM to simultaneously process both the visual and auditory content of a given video and engage in conversation with humans.
\\ 
\indent $\bullet$ To empower LLMs with video understanding capability, we propose a multi-branch cross-modal pre-training framework to achieve both vision-language alignment and audio-language alignment. \\
\indent $\bullet$ We open-source the entire codebase for pre-training and fine-tuning as well as the model weights of all the variants of Video-LLaMA\footnote{\url{https://github.com/DAMO-NLP-SG/Video-LLaMA}}. We also prepared the demos for video-grounded conversation\footnote{\url{https://huggingface.co/spaces/DAMO-NLP-SG/Video-LLaMA}}\footnote{\url{https://modelscope.cn/studios/damo/video-llama/summary}}.

\section{Method}
Video-LLaMA aims to empower frozen LLMs with the capability of understanding both visual and auditory content in videos. As shown in Figure~\ref{fig:architecture}, we design two branches, namely Vision-Language Branch and Audio-Language Branch, to respectively transform the video frames and audio signals into query representations that are compatible with the textual inputs of LLMs. In this section, we first introduce the overall architecture and the building blocks of each branch. Then, we delineate the procedures of the proposed multi-branch cross-modal pre-training and audio-visual instruction tuning.

\subsection{Architecture}
\subsubsection{Vision-Language Branch} 
The Vision-Language Branch is designed for enabling the LLMs to understand visual inputs. As shown in the left part of Figure~\ref{fig:architecture}, it is composed of a frozen pre-trained image encoder to extract features from video frames, a position embedding layer to inject temporal information into video frames, a video Q-former to aggregate frame-level representations and a linear layer to project the output video representations into the same dimension as the text embeddings of LLMs. Given one video consists of $N$ frames, the image encoder will first map each frame/image into $K_f$ image embedding vectors, yielding video frame representations $\mathrm{\bf V} = [\mathrm{\bf v}_1,\mathrm{\bf v}_2,...,\mathrm{\bf v}_N]$ where $\mathrm{\bf v}_i \in \mathbb{R}^{ K_f \times d_f}$ is the set of $d_f$-dimensional image embeddings  corresponding to the $i$-th frame.  

Since the frame representations $\mathrm{\bf v}_i$ from the frozen image encoder are computed without considering any temporal information, we further apply position embeddings as the indicator of temporal information to the representations from different frames. Then, we feed the position-encoded frame representations to Video Q-former, which shares the same architecture with Query Transformer (Q-Former) in BLIP-2~\citep{li2023blip2bl}, to obtain $k_V$ video embedding vectors of dimension $d_v$ as the representation $\hat{\mathrm{\bf v}} \in \mathbb{R} ^ {k_V \times d_v}$ of the video.

To adapt the video representations to the input of LLMs, we add a linear layer to transform the video embedding vectors into the video query vectors. The video query vectors are of the same dimension as the text embeddings of LLMs. In the forward pass, they will be concatenated to text embeddings as a \textit{video soft prompt} and guide the frozen LLMs to generate text conditioned on video content.

As for the implementation of the Vision-Language Branch, we utilize the pre-trained vision component of BLIP-2~\citep{li2023blip2bl} as the frozen visual encoder, which includes a ViT-G/14 from EVA-CLIP~\citep{fang2022eva} and a pre-trained Q-former. The remaining components, including the position embedding layer, Video Q-former, and Linear layer are randomly initialized and optimized to well connect the output of the frozen visual encoder to frozen LLMs.

\subsubsection{Audio-Language Branch}
To deal with the auditory content of the given video, we introduce the Audio-Language Branch. Concretely, it consists of a pre-trained audio encoder to compute features given a short segment of origin audio, a position embedding layer to inject temporal information to audio segments, an audio Q-former to fuse the features of different audio segments, and a linear layer to map the audio representation into the embedding space of LLMs.

In practice, we utilize the pre-trained Imagebind~\citep{girdhar2023imagebind} as the audio encoder. We first uniformly sample $M$ segments of 2-second short audio clips from the video, then convert each 2-second audio clip into spectrograms using 128 mel-spectrogram bins. After obtaining the spectrogram list of input audio, the audio encoder will map each spectrogram into a dense vector. So the generated audio representation of the given video can be denoted as  $A = [a_1,a_2,...,a_M]$. 

Similar to Video Q-Former, the Audio Q-former injects temporal information by adding learnable positional embeddings to audio segments. It then generates fixed-length audio features by computing the interaction across the position-encoded audio segments.  Audio Q-Former adopts the same architecture as Q-Former. It projects the variable-length audio representation list $A$ into a fixed-length sequence $\hat{\mathrm{\bf A}} \in \mathbb{R} ^ {K_a \times d_a}$, where the $K_a$ is the number of audio embedding vectors and $d_a$ is the dimension of each vector. Finally, we employ a linear layer to map audio features to the embedding space of the LLM.


\subsection{Multi-branch Cross-Modal Training}
We train the vision-language and audio-language branches separately. In the first stage, large-scale vision-caption datasets are used for training, and in the second stage, high-quality instruction-following datasets were used for fine-tuning. The image is treated as a one-frame video.

\subsubsection{Training of Vision-Language Branch}
For pre-training vision-language branch, we utilized Webvid-2M~\citep{Bain21}, a large-scale dataset of short videos with textual descriptions sourced from stock footage sites. Moreover, we employed the image caption dataset CC595k, which is sourced from CC3M~\citep{sharma-etal-2018-conceptual} and filtered by \citet{liu2023visualit}. We adopt a video-to-text generation task during the pre-training stage, i.e., given the representation of a video, prompting the frozen LLM to generate the corresponding text description. We find that a significant portion of textual descriptions are insufficient to reflect the entire content of the videos. Therefore, the visual semantics in the videos are not fully aligned with the textual semantics in the video descriptions. Nevertheless, this stage aimed to utilize a vast amount of data and enable video features to contain as much visual knowledge as possible. We left the abilities of vision-text alignment and instruction-following for the next stage.

After the  pre-training stage, the model can generate content about information in the video, but its ability to follow instructions has decreased. Therefore, in the second stage, we fine-tuned the model using high-quality instruction data. We integrated the image-detail-description dataset from MiniGPT-4~\citep{zhu2023minigpt}, the image-instruction dataset from LLaVA~\citep{liu2023visualit}, and the video-instruction dataset from Video-Chat~\citep{li2023videochatcv}. After fine-tuning, Video-LLaMA exhibited remarkable abilities in following instructions and comprehending images and videos. 

\subsubsection{Training of Audio-Language Branch}
Training the audio-language branch directly using audio-text data is highly challenging due to the rarity of such data.  The objective of the learnable parameters in the audio-language branch is to align the output embedding of the frozen audio encoder with the embedding space of LLM. Given the scarcity of audio-text data, we employ a workaround strategy to achieve this objective.
ImageBind, which is used as our audio encoder, has a remarkable ability to align different modalities' embeddings to one common space, demonstrating impressive performance on cross-modal retrieval and generation tasks. 
In light of the scarcity of audio-text data and the abundance of visual-text data, we train the audio-language branch using visual-text data, following the same data and process as the vision branch.  Thanks to the shared embedding space provided by ImageBind, Video-LLaMA exhibits the ability to comprehend audio during inference, even though the audio interface has never been trained on audio data.

\section{Related Works}
\textbf{Large Language Models}: Large language models (LLMs)~\citep{black2022gpt, scao2022bloom, openai2023gpt4tr, tsimpoukelli2021multimodal} have demonstrated remarkable language understanding and reasoning abilities, enabling the generation of high-quality natural language text across various domains, including articles, conversations, stories, and poetry. LLMs have already sparked a technological revolution and have been widely applied in different applications. Moreover, a series of open source large models, such as LLaMA~\citep{touvron2023llama}, BLOOM~\citep{scao2022bloom} and OPT~\citep{zhang2022opt}, have greatly promoted technological advancement and made outstanding contributions to the NLP community. Building upon these LLMs, researchers have further extended their capabilities and developed excellent models for various NLP tasks. Examples include Vicuna~\citep{chiang2023vicuna} and Baize~\citep{xu2023baize}. Our work is based on these LLMs and provides plug-and-play plugins that empower them with the capability of comprehending both visual and auditory content in videos.

\textbf{Multi-modal Large Language Models}: 
Researchers have been actively exploring the use of LLMs for processing multi-modal inputs~\citep{gao2023llamaadaptervp,li2023videochatcv}. Existing approaches can be categorized into two main groups. The first category involves employing LLMs as controllers and utilizing existing multi-modal models as tools. In this approach, when receiving the user's text instruction, the LLM recognizes the user's intention and makes decisions about which tools to call. It then generates comprehensive responses by incorporating the results obtained from these off-the-shelf multi-modal models. Examples include ChatGPT~\citep{wu2023visual}, HuggingGPT~\citep{shen2023hugginggpt}, and AudioGPT~\citep{huang2023audiogpt}. 
The second category focuses on training fundamental large-scale multi-modal models. The key idea of this line of work is to align the pre-trained foundation models for other modalities to textual LLMs. For instance, Flamingo~\citep{alayrac2022flamingo} utilizes a perceiver resampler and a gated cross-attention layer to connect a frozen image encoder and LLM.
BLIP2~\citep{li2023blip2bl} introduces a Q-Former to map learned image queries to the textual embedding space of LLMs. ~\citep{liu2023visualit}, mPLUG-owl~\citep{ye2023mplugowl} and MiniGPT4 ~\citep{zhu2023minigpt} develop instruction-following image-LLMs using image-instruction-following dataset. Video-Chat~\citep{li2023videochatcv} and Video-ChatGPT~\citep{maaz2023videochatgpt} extend image encoders to video encoders and connect them with LLMs to understand visual content in videos. PandaGPT~\citep{pandagpt} utilizes multi-modal encoders from ImageBind, trained exclusively on image-instruction pairs, to enable large models to understand six modalities. Our work falls into the second category, where we train fundamental models to comprehend both the visual and auditory content in videos.

\section{Examples}
In this section, we show some cases to demonstrate Video-LLaMA's  multi-modal instruction-following capability in video/audio/image-grounded conversations. Figure~\ref{fig.demo_case} shows some of these examples, while additional cases can be found in  Appendix~\ref{sec:appendix}.

\begin{figure*}[t]
    \centering
    \vspace{-0.5cm}
    \subfigure[A case that Video-LLaMA answers the questions based on the background sound and visual content of
the video]{
        \centering
        \includegraphics[width =0.45\linewidth]{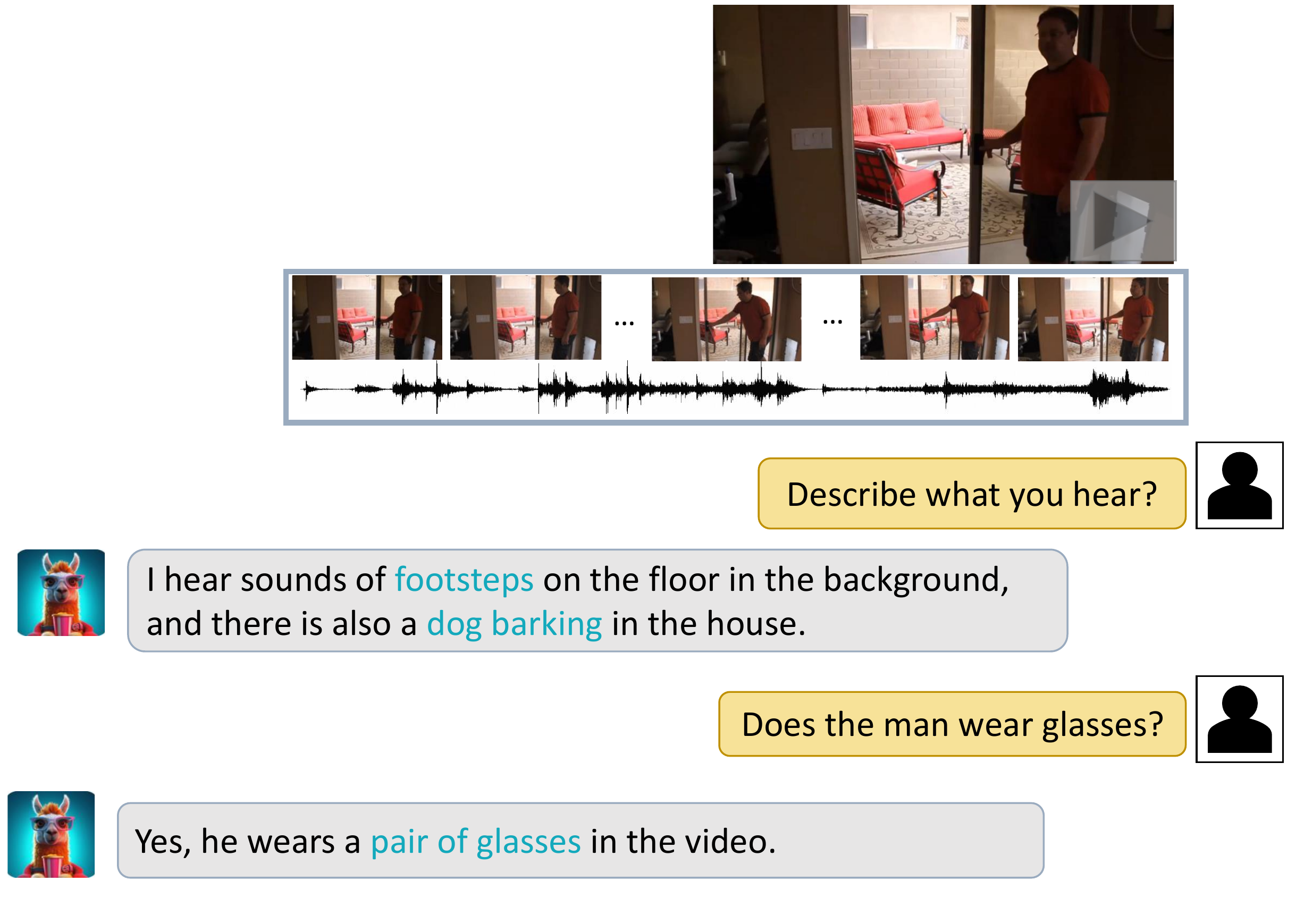}
        \label{fig.demo_case_audio_video_description}
        }
        \quad
     \subfigure[A case showing Video-LLaMA’s ability to identify actions over time.]{
        \centering
        \includegraphics[width =0.45\linewidth]{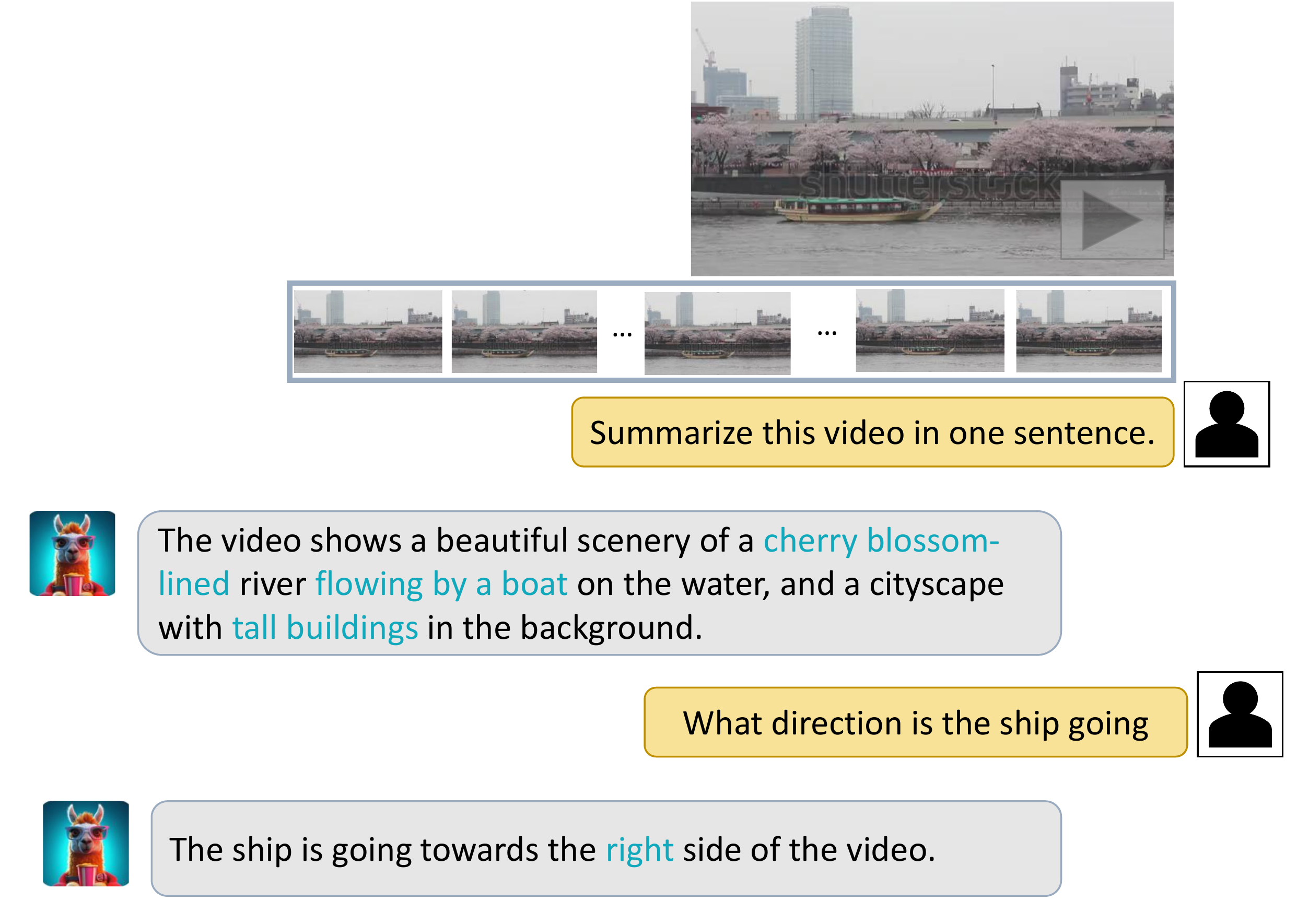}
        \label{fig.demo_case_video_qa}
        }

    \subfigure[A case demonstrating Video-LLaMA’s ability to comprehend static images.]{
        \centering
        \includegraphics[width =0.45\linewidth]{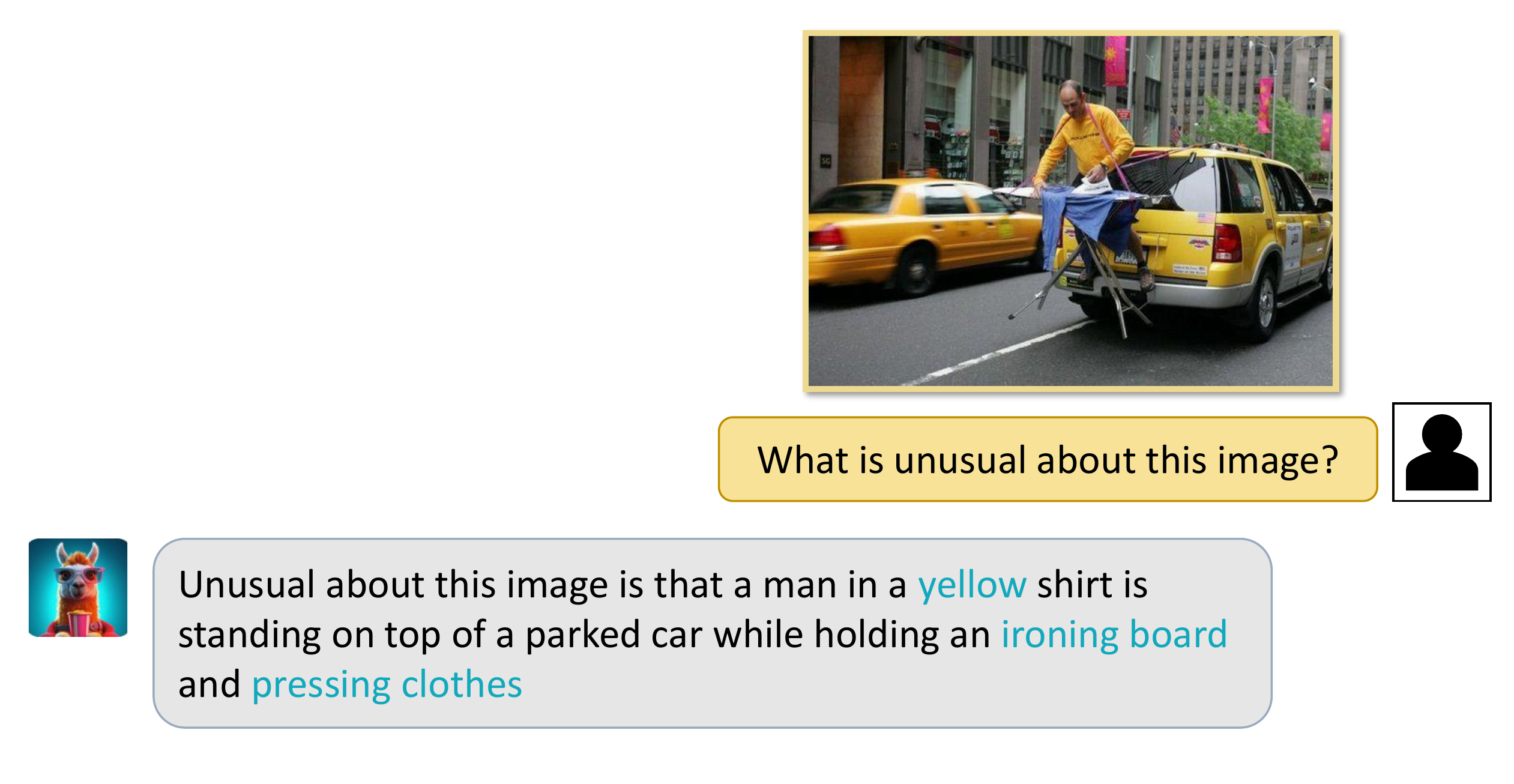}
        \label{fig.demo_case_image_understanding}
        }
    \quad
    \subfigure[A case demonstrating Video-LLaMA’s ability to recognize famous landmarks.]{
        \centering
        \includegraphics[width =0.45\linewidth]{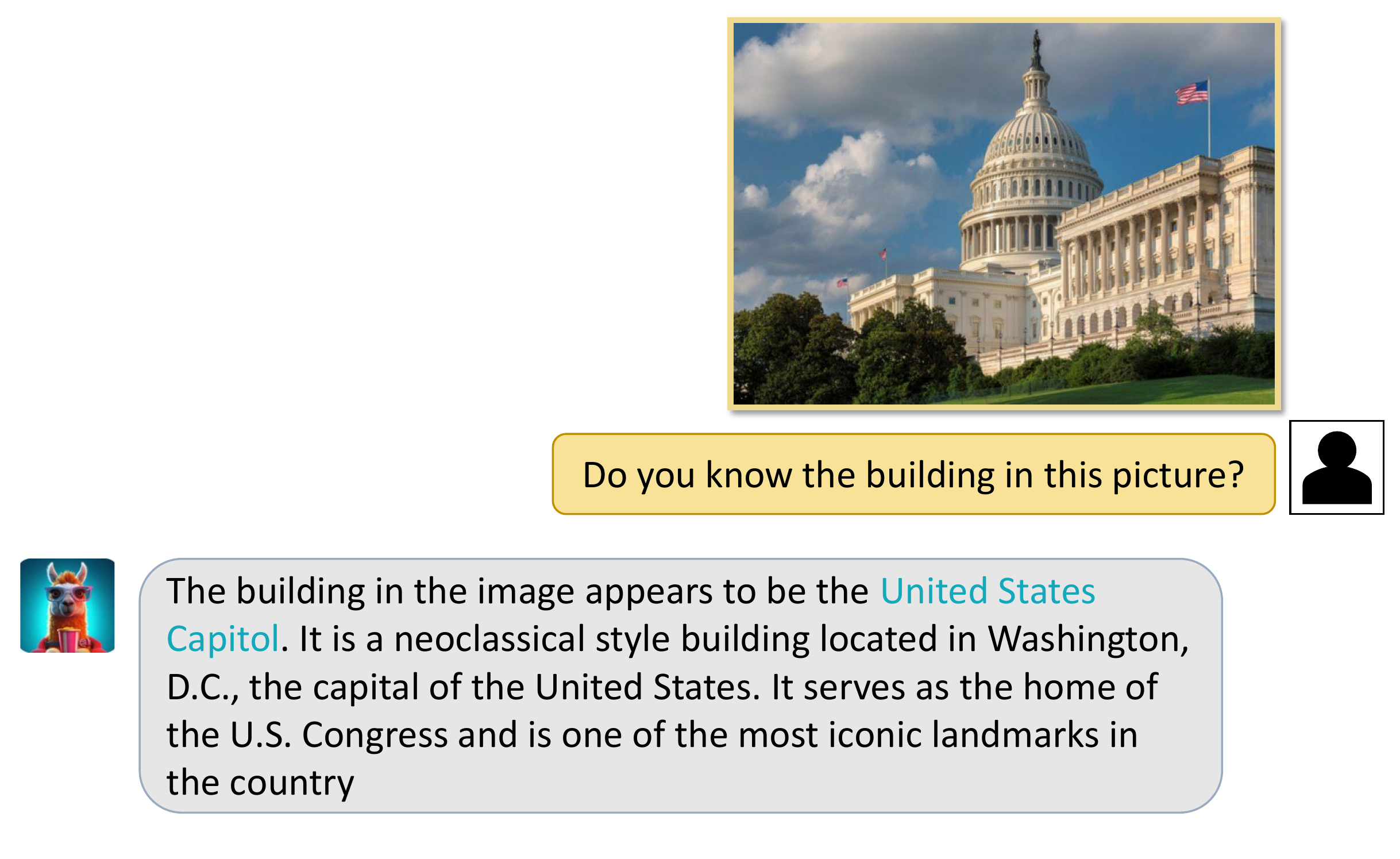}
         \label{fig.demo_case_image_concept_knowledge}
        }
\caption{Some examples generated by Video-LLaMA. }
    \label{fig.demo_case}
    \vspace{-0.3cm}
\end{figure*}


\paragraph{(1) Audio-visual integration perception ability.} Figure~\ref{fig.demo_case_audio_video_description} and Figure~\ref{fig.demo_case_audio_video_des_2} show Video-LLaMA's unique ability to comprehend auditory and visual information simultaneously. The videos in both cases contain audio. In each conversation, we pose two questions related to visual and auditory content respectively.  If the model could only receive one modal, it would be unable to answer both of these questions. However, we can observe that Video-LLaMA accurately responds to both visual and auditory questions in both cases.

\paragraph{(2) The ability to capture temporal dynamics in videos.} 
Figure~\ref{fig.demo_case_video_qa} and Figure~\ref{fig.demo_case_video_description} illustrate the capability of Video-LLaMA to identify actions over time. 
It successfully describes the actions of the girl and the moving direction of the boat.

\paragraph{(3) The ability to perceive and understand static images.} Figure~\ref{fig.demo_case_image_understanding} and Figure~\ref{fig.demo_case_image_des}  show Video-LLaMA's ability to perceive and understand pictures. Figure~\ref{fig.demo_case_image_understanding} demonstrates Video-LLaMA's ability to understand the concept of "unusual" and specifically describe the unusual scene.  In Figure~\ref{fig.demo_case_image_des}, not only does Video-LLaMA accurately describe the main content, but it also associates it with the friendly interaction between a dog and a human.

\paragraph{(4) The ability of common-knowledge concept recognition.} Figure~\ref{fig.demo_case_image_concept_knowledge} and Figure~\ref{fig.demo_case_image_multiturn} demonstrate Video-LLaMA's remarkable capacity for recognizing common-knowledge concepts in visual signals. Video-LLaMA successfully recognizes famous landmarks and characters and can engage in common-sense question-answering.

\section{Conclusion}
In this paper, we present Video-LLaMA, a cutting-edge multi-modal framework that empowers large language models with both audio \& video understanding capabilities. Our experiments  demonstrated the impressive abilities of Video-LLaMA in audio and video-grounded conversations, highlighting its potential as a promising prototype for audio-visual AI assistants. We have open-sourced the entire training code and various model weights, along with detailed instructions to assist developers in utilizing our code for further development. In addition, we have provided online demo websites and offline demo deployment guides for users to experience Video-LLaMA's capabilities firsthand. We are committed to constantly maintaining and improving Video-LLaMA, and will continue to contribute to the open-source community.

\section{Limitations}
Although Video-LLaMA has demonstrated impressive abilities in understanding both visual and auditory content in videos, it is still an early-stage prototype and has some limitations, including:
(1) Limited perception capacities: Video-LLaMA's performance is hindered by the quality and scale of the current training dataset. We are actively constructing a high-quality audio-video-text alignment dataset to enhance the model's perception capabilities. 
(2) Limited ability to handle long videos. Long videos(such as movies, and TV shows) contain a large volume of information and impose higher demands on computational resources. This challenge remains a crucial issue that the research community is actively working to address.
(3) Hallucination. Video-LLaMA inherits the hallucination problem from the frozen LLMs. We will continue to address these challenges and develop more powerful versions for video understanding.

\clearpage
\bibliography{anthology,custom}
\bibliographystyle{acl_natbib}

\appendix

\section{Appendix}
\label{sec:appendix}

\begin{figure*}[ht]
    \centering
    \includegraphics[scale=0.44]{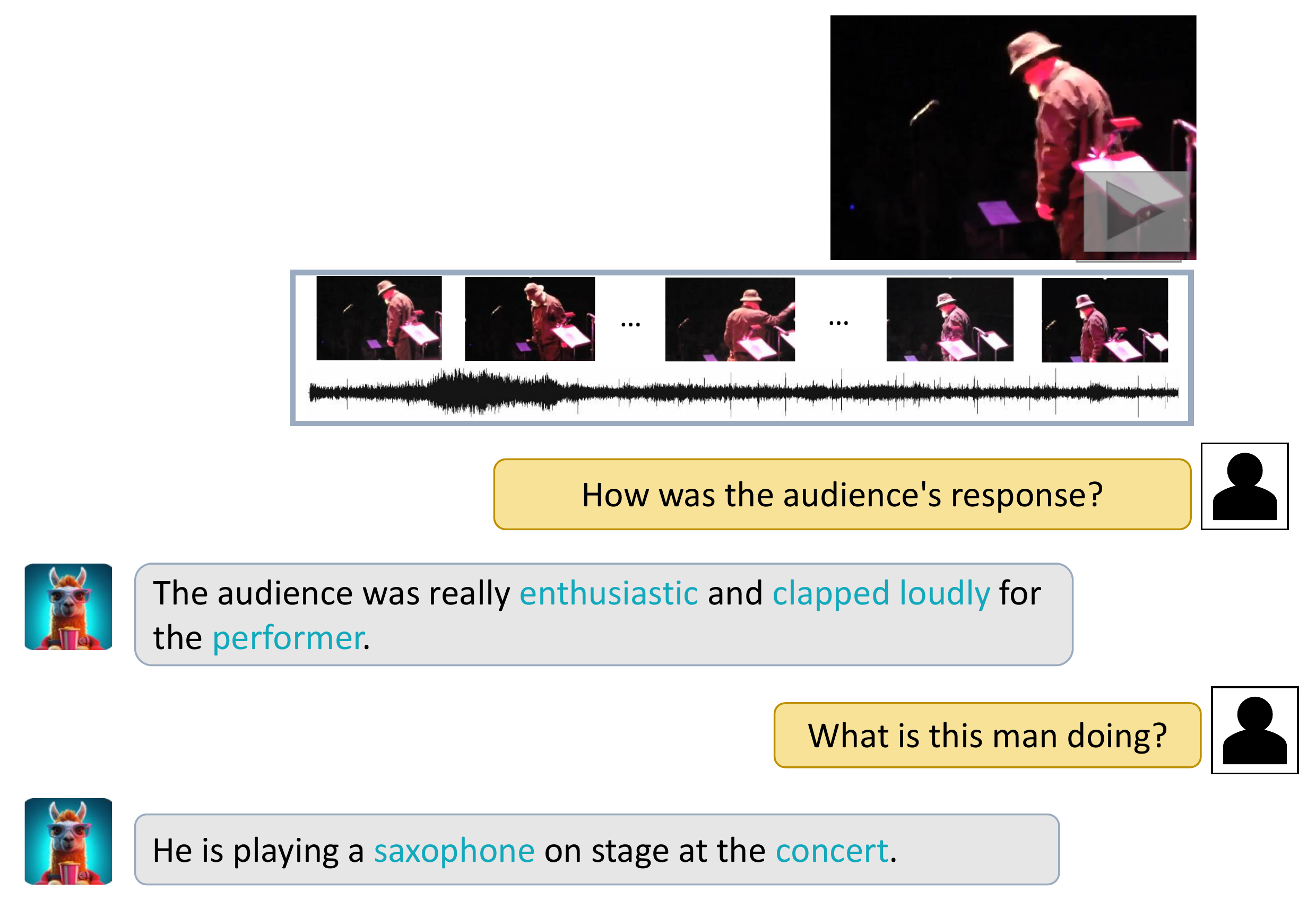}
    \caption{A case showing Video-LLaMA's ability to identify the sound of applause in a video and infer the positive response from the audience. Additionally, it infers that a man is playing the saxophone on stage based on the visual content.}
    \label{fig.demo_case_audio_video_des_2}
\end{figure*}

\begin{figure*}[ht]
    \centering
    \includegraphics[scale=0.44]{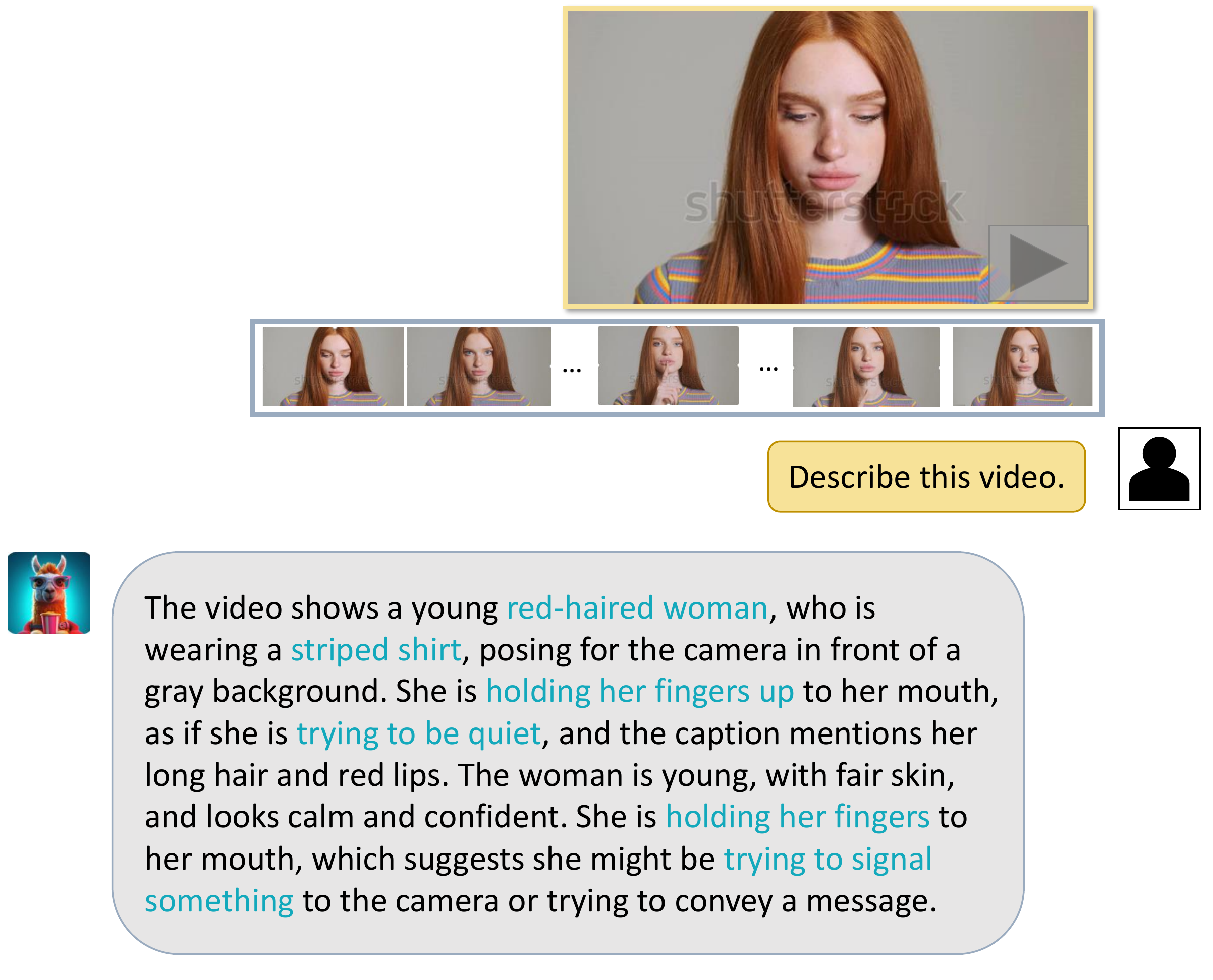}
    \caption{A case where Video-LLaMA provides a detailed description of the visual content in a dynamic video.}
    \label{fig.demo_case_video_description}
\end{figure*}

\begin{figure*}[ht]
    \centering
    \includegraphics[scale=0.44]{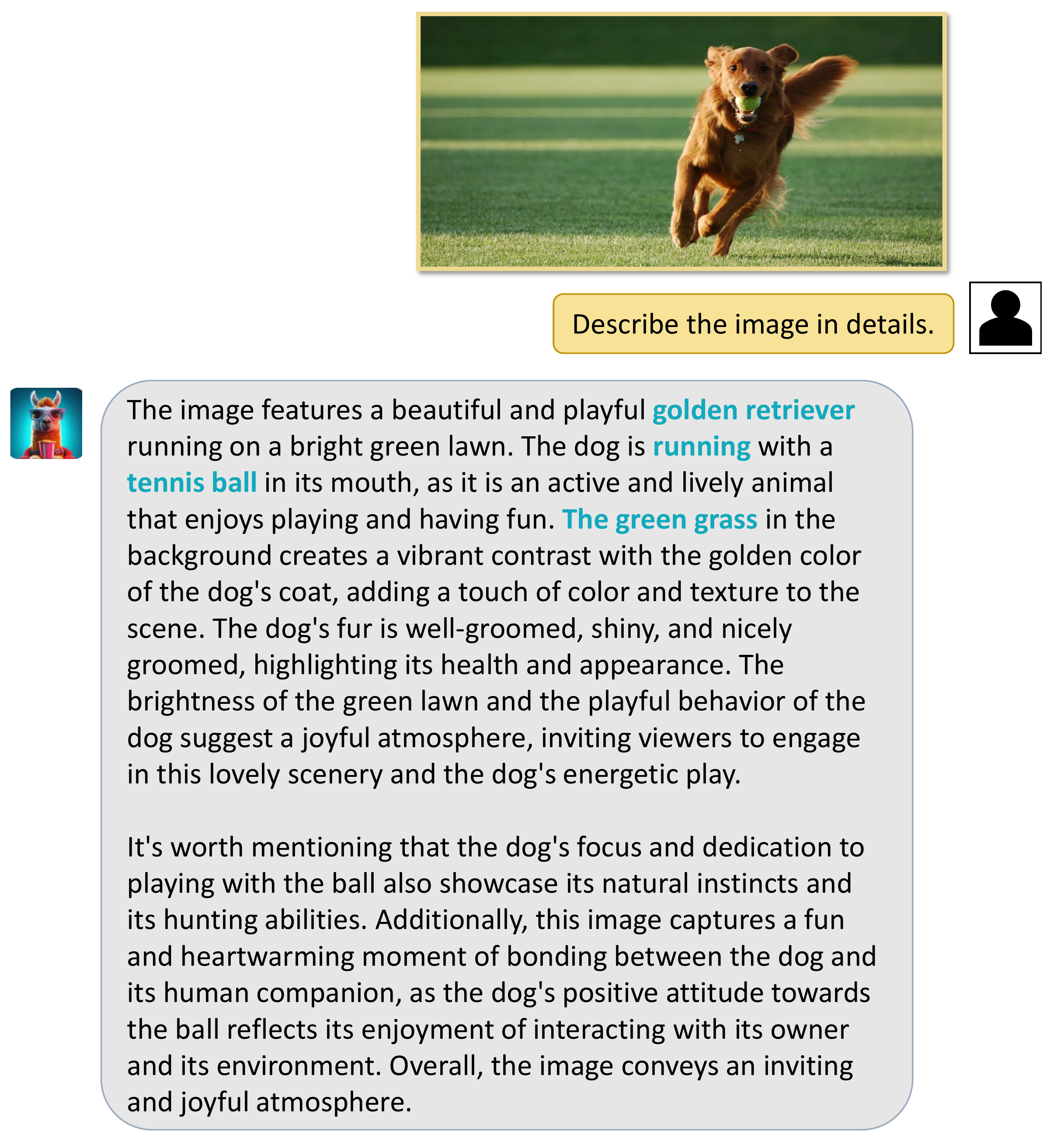}
    \caption{A case where Video-LLaMA provides a detailed description of the static image content.}
    \label{fig.demo_case_image_des}
\end{figure*}



\begin{figure*}[ht]
    \centering
    \includegraphics[scale=0.44]{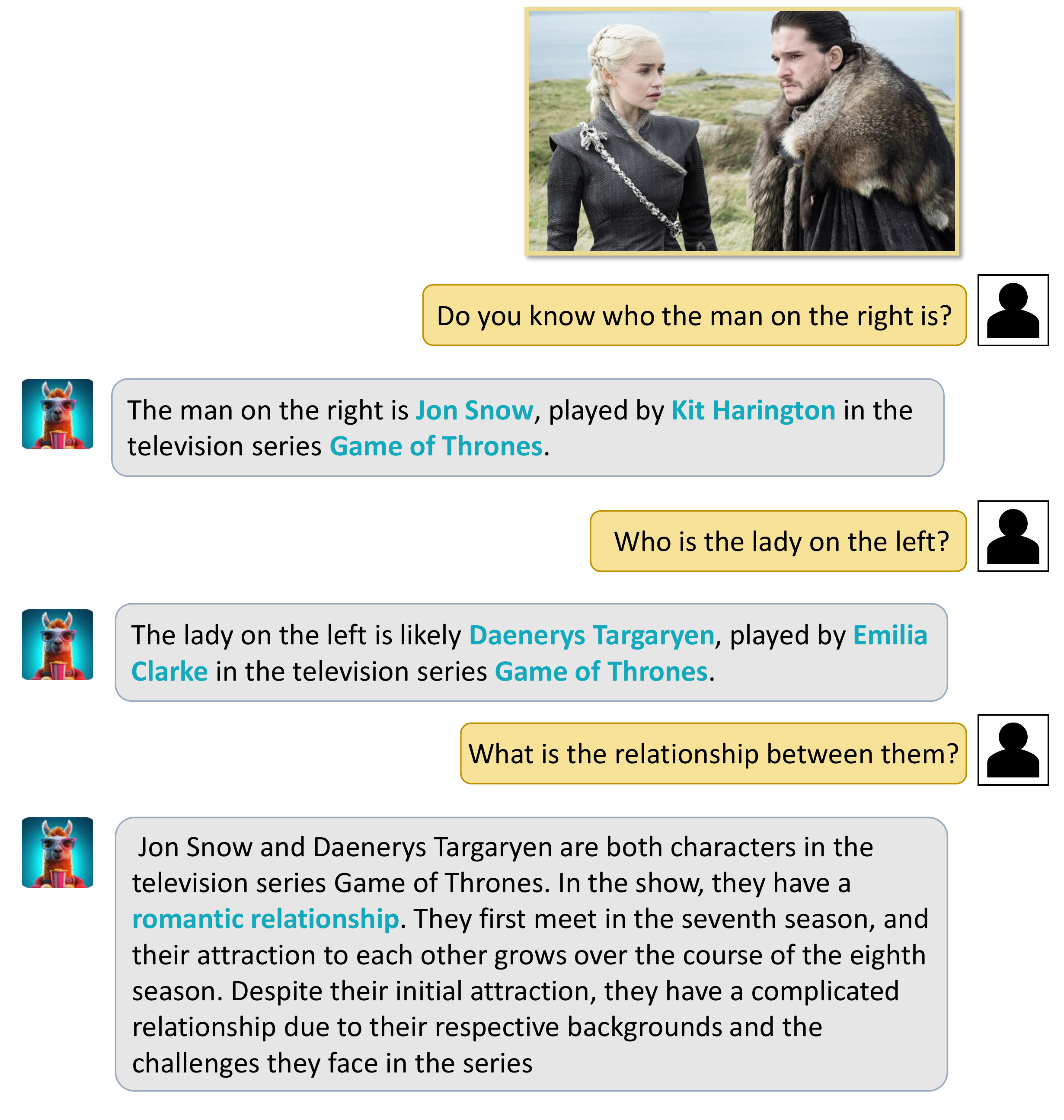}
    \caption{A case showing Video-LLaMA's ability to recognize renowned characters and participate in video-grounded question answering.}
    \label{fig.demo_case_image_multiturn}
\end{figure*}

\end{document}